\documentclass[conference]{IEEEtran}
\IEEEoverridecommandlockouts

\usepackage{cite}
\usepackage{amsmath,amssymb,amsfonts,amsthm,mathtools}
\usepackage{algorithmic}
\usepackage{graphicx}
\usepackage{textcomp}
\usepackage{xcolor}
\usepackage{comment}
\usepackage{color,soul}

\def\BibTeX{{\rm B\kern-.05em{\sc i\kern-.025em b}\kern-.08em
    T\kern-.1667em\lower.7ex\hbox{E}\kern-.125emX}}
\begin{document}

\title{Quality of Control based Resource Dimensioning for Collaborative Edge Robotics
}

\author{\IEEEauthorblockN{Neelabhro Roy, Mani H. Dhullipalla, Gourav Prateek Sharma, Dimos V. Dimarogonas \& James Gross}
\IEEEauthorblockA{ \text{School of Electrical Engineering and Computer Science \& Digital Futures Research Center} \\
\text{KTH Royal Institute of Technology}
Stockholm, Sweden \\
\{nroy, manihd, gpsharma, dimos, jamesgr\}@kth.se}
}

\maketitle

\begin{abstract}
With the increasing focus on flexible automation, which emphasizes systems capable of adapting to varied tasks and conditions, exploring future deployments of cloud and edge-based network infrastructures in robotic systems becomes crucial. This work, examines how wireless solutions could support the shift from rigid, wired setups toward more adaptive, flexible automation in industrial environments. We provide a quality of control (QoC) based abstraction for robotic workloads, parameterized on loop latency and reliability, and jointly optimize system performance.
The setup involves collaborative robots working on distributed tasks, underscoring how wireless communication can enable more dynamic coordination in flexible automation systems. We use our abstraction to optimally maximize the QoC ensuring efficient operation even under varying network conditions. Additionally, our solution allocates the communication resources in time slots, optimizing the balance between communication and control costs. 
Our simulation results highlight that minimizing the delay in the system may not always ensure the best QoC but can lead to substantial gains in QoC if delays are sometimes relaxed, allowing more packets to be delivered reliably.
\end{abstract}

\begin{IEEEkeywords}
collaborative robotics, safety-critical applications, quality of control, multi-agent systems, edge computing
\end{IEEEkeywords}

\section{Introduction}
Flexible automation and mass customization are key drivers in this era of smart manufacturing and industrial development \cite{5gsmart2020}, contributing to advancing the goals of Industry 4.0 and 5.0. They can enable systems to adapt to varying production needs with minimal reconfiguration, making handling diverse and complex tasks easier, and finally paving the way for producing personalized products at scale.
With the emergence of flexible automation, revisiting the mostly wired robotic solutions prevalent in the industry is required, as they do not offer mobility, adaptability and restrict collaboration. For such joint tasks, wireless collaborative robotics is crucial~\cite{7883994}.

Wireless collaborative robotics necessitate managed systems promising low latency, high reliability, and flexibility, where 5G networks in combination with local edge computing seem attractive \cite{edge_robotics}. These 5G-based systems offer the possibility to offload different computational tasks to the edge cloud.
However, practical implementations of collaborative robotics \cite{mesbahi2010graph} where communications and compute are provided by 5G and edge computing-based infrastructures, respectively, are not well understood, opening some questions:
(a) What are the different information flows in a collaborative robotic setup and how can their looped interactions be mapped to a 5G-based implementation?
(b) How would the robotic setup react to uncertainties in terms of delays, reliability, capacity constraints emanating from a 5G setup?
(c) Can an acceptable quality-of-control (QoC) \cite{QoC_Assessment} tracking the energy and control expenditure, be maintained? and finally,
(d) How to develop an abstraction of robotic workloads parameterized by these uncertainties, to jointly optimize system performance?

There have been a few works addressing these questions: Some works such as \cite{latency_min}, minimize the end-to-end latency for networked robotic applications like automated guided vehicles (AGVs) on a factory floor. In their work, provided a delay requirement for their system, they find an allocation satisfying that requirement with fixed reliability bounds. While such works minimize the end-to-end latency, they lack an understanding of network reliability and how that might affect the robotic system. 
\cite{collab4g, WifiEdge} provide orchestration frameworks for networked collaborative robotics that offload motion level control to the edge. They highlight the importance of minimizing the time for the successful completion of mission-critical tasks and assume packet error rate (PER) and delay thresholds, with \cite{collab4g} also indicating capacity constraints. 
All these works, however, have a binary value function, where as long as a certain latency, PER or task success rate target is met, the system is considered stable.
This calls for a joint understanding of latency and reliability to characterize the performance of the control application with a measurable metric such as QoC. 
Authors in \cite{QoC_Assessment} propose a QoC metric that is sensitive to control loop instabilities, estimating positional errors in tactile-visual control applications. While they individually show how QoC is affected by end-to-end latency, jitter, and packet drops, they don't provide a coherent abstraction of QoC based on their combined understanding for robotic workloads and still lack in indicating relevant tradeoffs and joint optimization potentials. \cite{lat-rel} tackles this by providing such an abstraction based on tradeoffs between latency and reliability for the problem of estimating dynamical systems over communication channels. Using this abstraction they obtain an optimal code-block length for the state estimation problem. Their communication model, however, is based on information/coding theory for finite blocklengths.
Overall, we observe a dearth in literature where an abstraction of a collaborative robotic system parameterized with both latency and reliability is used to jointly optimize control applications connected with a 5G network. We aim to fill this gap, assuming a multi-robot setup where they collaborate towards a joint task, having offloaded their motion-level control to the 5G edge, with our contributions:
\begin{itemize}
    \item We derive a QoC-based abstraction for collaborative robotics, considering both network delay and reliability.
    \item We exemplify the applicability of our approach with a consensus-based scenario, by considering a joint robotic system and network optimization. 
    \item Our simulation results show that by relaxing the delays in some cases, we save up to 32\% in energy consumption as compared to State-of-the-art (SOTA) evaluation schemes.
\end{itemize}
The rest of the paper is structured as follows:
In Section~\ref{system model} we introduce the system model before we derive QoC for a collaborative robotic system in Section~\ref{QOC_derivation}.
We present our optimization framework, evaluation methodology and results in Section~\ref{results}, while the paper is concluded in Section~\ref{conclusion}.

\section{System Model and Problem Statement} \label{system model}
In this section, we first describe the system model with a focus on collaborative robotics, and a supporting communication and computation model. Subsequently, we state the problem that this work aims to address. 
\subsection{Collaborative Robotics}
\begin{figure}[t]
    \centering
    \includegraphics[width=0.99\linewidth]{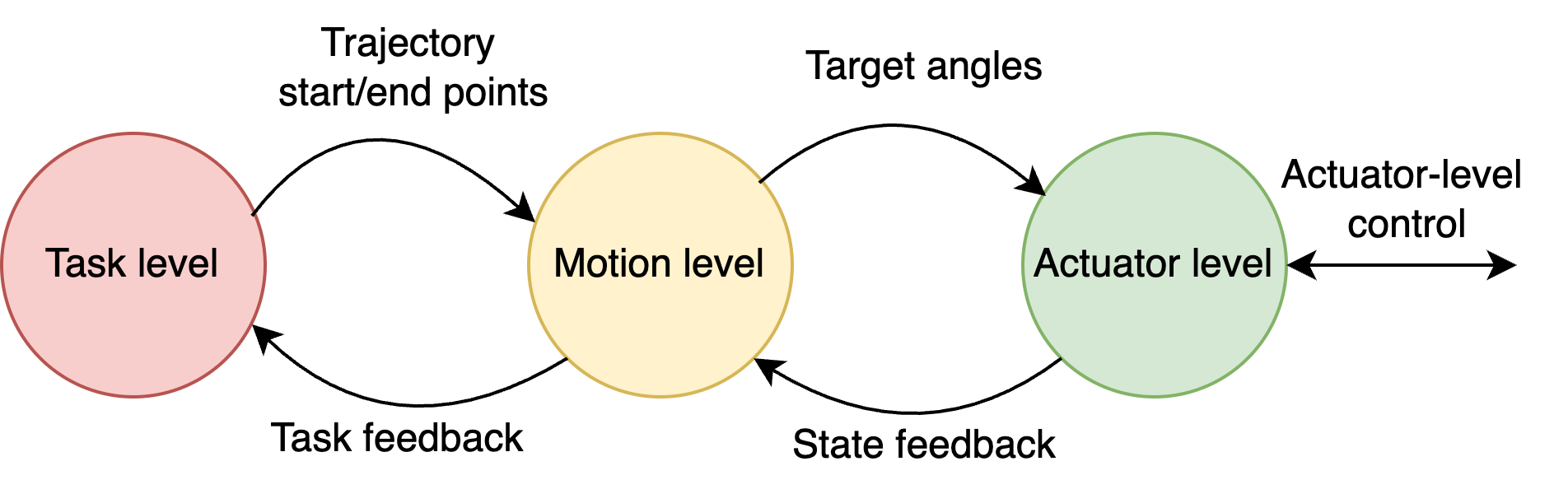}
    \caption{Different levels of control}\label{fig:control_level}
\end{figure}

Collaborative robotics, as the name suggests, involves coordination amongst multiple robots to collectively accomplish tasks or goals such as surveillance, formation control, and optimal coordination, see \cite{cao2012overview}. Since coordination amongst robots is crucial for task accomplishment, based on the complexity of tasks, the information flow in multi-robot systems may broadly be classified into the following levels of control (see Fig. \ref{fig:control_level}):

\begin{itemize}
    \item \textbf{Task level}: This level involves information exchanges necessary for long-term task planning. For instance, the central server and/or robots exchange information regarding target positions every few seconds, taking the real-time status of the overall task into account. 
    \item \textbf{Motion level}: To achieve the target poses, the robots splice the target information to the second level which involves motion-planning from their current pose to the target pose. To safely manoeuvre in the available space, the robots frequently exchange their position with each other; these exchanges happen every 4 - 10ms \cite{s21227463}. 
    \item \textbf{Actuator level}:  Finally, the third level comprises actuator control that facilitates motion in the robots by applying forces/torques based on information from the second level, typically operating in the 100$\mu$s - $1$ms range \cite{10558844}.
\end{itemize}

From a communication standpoint, the level of control most crucial for collaborative robotics is motion level, since the robots exchange their position (or, more generally, state) information to achieve a joint objective such as rendezvous, formation control, optimal coordination \cite{cao2012overview, mesbahi2010graph}. Once the state exchange is done, appropriate control commands can be applied. This loop of data exchange is crucial for system stability and depends on the underlying communication network.

To study the effect that communication networks have on collaborative robotics, we consider the problem of consensus, described more formally in Section \ref{QOC_derivation}. Consensus is one of the fundamental problems in multi-robot coordination and distributed sensor estimation, see \cite{mesbahi2010graph}. In robotics, this translates to the problem where multiple robots/agents exchange their state (or position) information and gather at some location (say, meeting point). Note that the meeting point is often not known \textit{a priori} and, therefore, frequent information exchange among the robots is necessary. Furthermore, the convergence rate (towards the meeting point) depends on: i) periodicity of information exchange, ii) latency and reliability of the communication networks, iii) overall number of robots involved, iv) the topology of how the robots are connected, and v) retransmissions / errors in the communication network. Thus, the robots have varying amounts of control expenditure, quantified using QoC defined in Section \ref{QoC_model}.

\begin{figure}[t]
    \centering
    \includegraphics[width=0.99\linewidth]{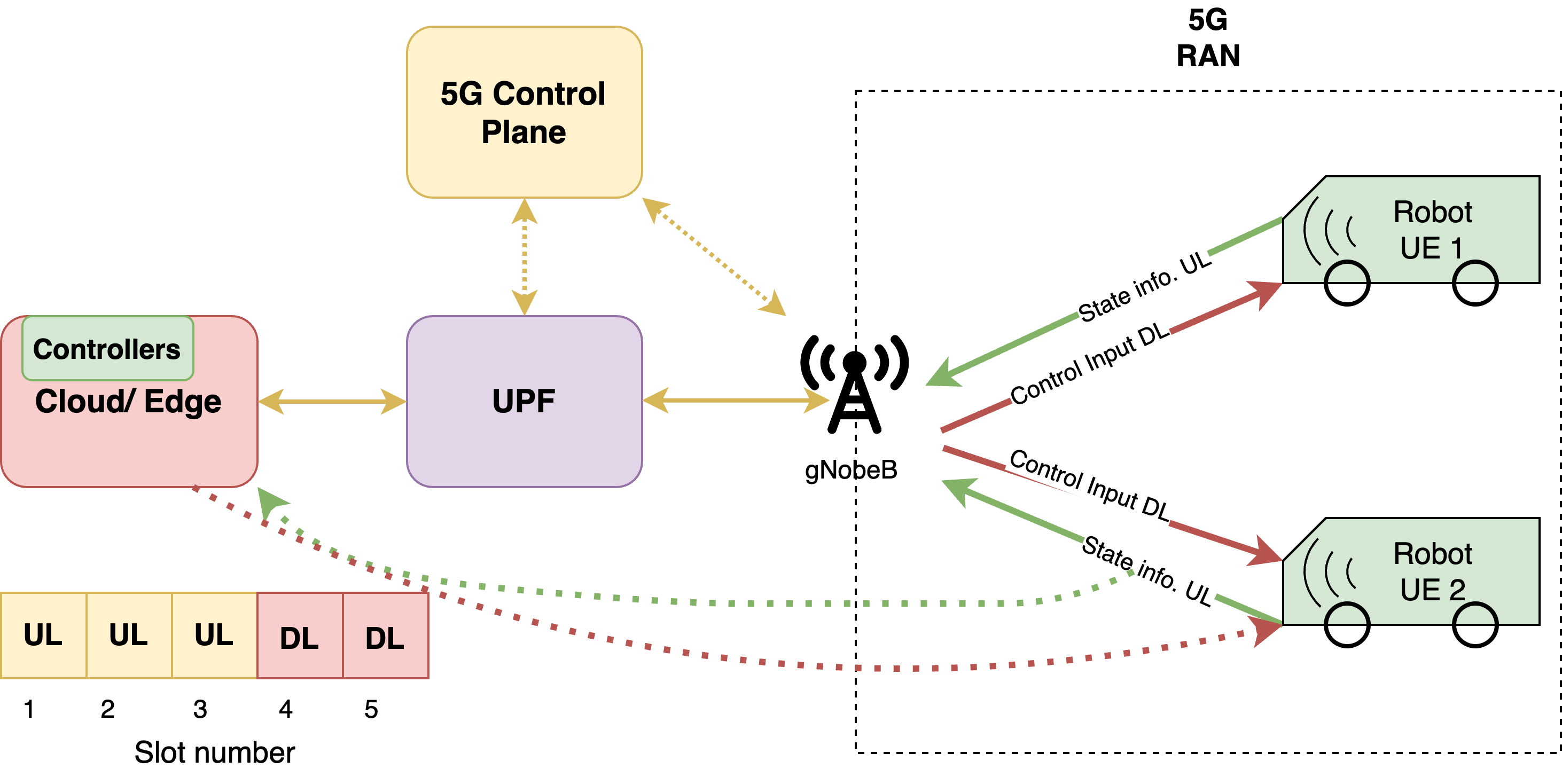}
    \caption{Overview of the system architecture with the motion-level control offloaded to the edge/cloud.}\label{fig:full_off}
\end{figure}
\subsection{Communication and Computation}
Now, we detail the communication model, which facilitates the loop data exchange described previously. The state information exchange and their corresponding control commands act as the system load, which needs to be transmitted at a fixed periodicity. To facilitate this, we assume a 5G system with edge computing infrastructure, as shown in Fig. \ref{fig:full_off}.
The first part of the 5G system is the radio access network (RAN) where the robots are physically located and connected to the gNodeB. This is where the information loop for a robot $i$ begins, with state information being directed uplink to the core network comprising the user plane function (UPF). UPF further routes this information to the edge and the 5G control plane. This transmission can be erroneous, with the PER being denoted with $\text{PER}_{\text{UL}}$. Upon reception of the information at the edge cloud running the controllers for the robots, the corresponding control commands for the next time step are computed and sent. This control computation coupled with random processing times at the core, however, can delay the packet further, randomly.
We can model these delays with independent and identically distributed (IID) random variables $D_{\text{back}}(t)$ whose probability density function (PDF) is denoted by $p_{\text{back}}$ and with a cumulative distribution function (CDF) denoted by $F_{\text{back}}(d)$. Assuming $d_i^{\text{alloc}}$ to be the allocation delay a robot experiences at the core and edge, we can calculate the probability that the allocation delay $D$ for robot $i$ is less than or equal to $d_i^{\text{alloc}}$ with:
\begin{equation}
P(D \leq d_i^{\text{alloc}}) = F_{\text{back}}(d_i^{\text{alloc}}) \label{rel_asb}
\end{equation}
Now, the control packet is sent back to the robot in the RAN, thus completing the information loop. This downlink transmission can again be erroneous with a PER given by $\text{PER}_{\text{DL}}$.
We combine all the uncertainties and errors to calculate the final loop end-to-end reliability corresponding to delay $d_i^{\text{alloc}}$ with:
\begin{equation}
    P(\text{success}) = F_{\text{back}}(d_i^{\text{alloc}})*\text{PER}_{\text{UL}}*\text{PER}_{\text{DL}}
    \label{reliability}
\end{equation}
This expression finally provides us the end-to-end reliability measure dictating the likeliness of a control packet being delivered successfully and is used in Section \ref{QOC_derivation}.

\subsection{Problem Statement}
In this paper, we offload the motion-level control of $N$ collaborating robots to a 5G edge computing-based setup, making the looped information exchanges between any two robots 
susceptible to network latency and reliability-based uncertainty. The objective is to develop an abstraction for such a setting to understand system performance beyond the binary understanding of system stability \cite{latency_min,collab4g}. Towards this, a qualitative metric of QoC $Q_i,   \quad \forall i\in\{1,\cdots,N\}$ capturing convergence for a robot $i$ and parameterized by latency and reliability needs to be developed. Moreover, we wish to see if $Q_i$ can be used to jointly optimize system performance in a collaborative edge robotics based setup.
\section{QoC derivation for Collaborative Robotics} \label{QOC_derivation}
This section demonstrates how a QoC-based system abstraction, considering network delays and reliability, can optimize performance in consensus-based collaborative robotics. Fig. \ref{fig:full_off} shows AGVs trying to reach a common meeting point by exchanging control information over a 5G edge-compute network. Depending on the looped interaction of control information happening with a certain network delay and reliability, the AGVs may converge or diverge, with varying convergence rates. We now showcase how a QoC-based metric tracking stability and convergence for such systems can be developed.

\subsection{System Dynamics} \label{subsec:system dynamics}
In this subsection, we present a simplified model \cite{mesbahi2010graph} to describe the robot's motion and subsequently provide the control input that is necessary for the robots to rendezvous at a common meeting point. Consider a multi-robot system with $N$ robots, where each robot (say, $i$) employs the following dynamics that guide its motion: 
\begin{equation}
    \dot{x}_i(t) = u_i(t),  \quad i\in\{1,\cdots,N\}. 
\end{equation}
Here, $x_i$ denotes the position of the robot, $\dot{x}_i$ denotes its velocity, $u_i$ denotes the control input and $t$ denotes time. The multi-robot system is said to achieve consensus if as time $t \to \infty$, $ x_i(t) = x_C,$ for every robot $ i \in \{1,\cdots, N\}$, where $x_C$ denotes the consensus position (not known \textit{a priori}). 

In order to achieve consensus, consider the following control law for each robot $i \in \{1,\cdots,N\}$:
\begin{equation} \label{eq:ctrl_law}
    u_i(t) = \kappa \sum_{z=1}^N (x_z(t) - x_i(t)), 
\end{equation}
where $\kappa > 0$ is the control gain. From \eqref{eq:ctrl_law}, one can infer that the motion of robot $i$ (i.e., $\dot{x}_i = u_i$) at every time instant $t$ is influenced by the relative position of every robot  $z \in \{1,\cdots,N\}$ with respect to that of robot $i$.  

To implement the control law in \eqref{eq:ctrl_law} over communication networks, the robots' positions $x_i$-s need to be frequently sent to the 5G edge and the control input computed at the edge needs to be sent back to the robots. In this work, we consider that the robots push the information to the 5G edge every 10ms \cite{s21227463}. These transmissions (to-and-fro) from the 5G edge are inevitably prone to network latency and reliability concerns.  As a result, a simplified model of networked implementation of robot dynamics is as follows:
\begin{align} \label{eq:ctrl_comm}
    \dot{x}_i = \hat{u}_i = \kappa \sum_{z=1}^N (\hat{x}_z - \hat{x}_i), \quad i\in\{1,\cdots,N\}.
\end{align}
Here, $\hat{u}_i$ denotes the control input received by robot $i$ from the 5G edge. In order to compute this input, the edge utilizes the last successfully transmitted robots' position denoted by $\hat{x}_z,\; z\in\{1,\cdots,N\}$. Note that the notation $\hat{x}_i$ is used to differentiate the transmitted position from the actual position $x_i$ of the robot and is computed using \eqref{reliability}.

\subsection{Quality of Control} \label{QoC_model}
In this subsection, we introduce a metric called quality of control (QoC) to assess the effect of reliability and delay on the system performance. Traditionally, in the literature \cite{QoC_energy_2}, this measure is evaluated using an energy-like function $J$ involving states $x$ (for e.g., position of robots) and input $u$ given by:
\begin{equation} \label{eq:controlenergy}
    J = \frac{1}{2} \int_0^{T} (x^T W x + u^T P u)\; dt
\end{equation}
where $W$ and $P$ are positive-definite matrices and $T < \infty $ denotes the time horizon. 

In this work, we borrow the measure in \eqref{eq:controlenergy} to define QoC for the consensus problem discussed in \ref{subsec:system dynamics}. First, for all robots $i\in\{1,\cdots,N\}$, let 
\begin{equation}
\delta_i(t) = x_i(t) - x_C,
\end{equation}
denote the disagreement state associated with robot $i$. Here, $\delta_i$ represents the relative position of robot $i$ with respect to the consensus state $x_C$. Note that the consensus problem in states $x_i$ (i.e., $x_i \to x_C,\; i\in\{1,\cdots,N\}$) is equivalent to stabilization of $\delta_i$ (i.e., $\delta_i \to 0, \;i\in\{1,\cdots,N\}$). The dynamics of disagreement for robot $i$, is given by $\dot{\delta}_i = \dot{x}_i $ and the control input $u_i$ can be expressed in terms of disagreement $\delta_i$ as follows:
\begin{align}
    u_i &= \kappa \sum_{z=1}^N (x_z - x_i) = -\kappa \sum_{z=1}^N (x_i - x_C) \nonumber
    \\ &= -\kappa N(x_i - x_C) = -\kappa N\delta_i.
\end{align}
Therefore, the dynamics of disagreement of robot $i \in\{1,\dots,N\}$ is given by $\dot{\delta}_i = -\kappa N \delta_i $; furthermore, this continuous-time system decays to $\delta_i = 0$ (in other words, achieves consensus) exponentially fast.  


Now, to define QoC, we employ the disagreement $\delta_i$ attached to each robot $i\in\{1,\dots,N\}$ in the multi-robot system. Since both the state expenditure (say, $|x_i-x_C|^2 $) and control expenditure (say, $ |u_i|^2$) can be expressed in terms of disagreement expenditure $|\delta_i|^2$, we define $J_i^C$ for each robot $i\in\{1,\dots,N\} $ as follows: 
\begin{equation}
    J_i^C = \int_0^T |\delta_i(t)|^2 dt. \label{disagreement}
\end{equation}
Furthermore, we normalize this measure with respect to the initial configuration (namely, $\delta_i(0), \; i\in\{1,\dots,N\}$) of the multi-robot system as follows: 
\begin{equation} \label{eq:AuC_defn}
    J_{\text{norm}_i}^C = \int_0^T \frac{|\delta_i(t)|^2}{|\delta_i(0)|^2} dt.
\end{equation}
This quantity is referred to as the area under the disagreement curve (AUC) for robot $i$. 

Finally, in this work, QoC is defined as follows: 
\begin{equation}
    Q_i = (\max(J_{\text{norm}_i}^C) - J_{\text{norm}_i}^C )/\max(J_{\text{norm}_i}^C) \quad 
\end{equation}
Here, $\max(J_{\text{norm}_i}^C)$ depends on the evaluation scenario and is determined when the QoC table is computed, as detailed in Section \ref{sec:evaluation}.
The physical interpretation of $Q_i$ here can be understood with energy-based comparisons: twice the $Q_i$ on average for a system, as compared to another, would indicate half the energy expenditure in time $T$, achieving consensus.

\subsection{QoC-based System Abstraction} \label{QOC abstraction}
We begin this section by detailing how a QoC model could be obtained, given a robotic system. 
We want to observe how Eq. \eqref{disagreement} can help us understand the delay reliability tradeoff and can be used to track convergence towards consensus.
Towards this, we simulate a system with $N =$ 80 robots in a complete graph with $\tau = 10\text{ms}$ periodicity. $D_{\text{back}}(t)$ is chosen as a Gaussian distribution with a standard deviation of 1ms and a mean of 0.5ms, normalized and clipped between 0ms and $\tau = 10\text{ms}$. The reliability increases with increasing delays as Eq. \ref{rel_asb} indicates.
With the dynamics of this system from \eqref{eq:ctrl_comm}, we run 1000 simulations and average the results. The delays and their reliability measures are fixed for these simulations, with the random delays varying across simulation runs.

\begin{figure}[t]
    \centering
    \includegraphics[width=0.99\linewidth]{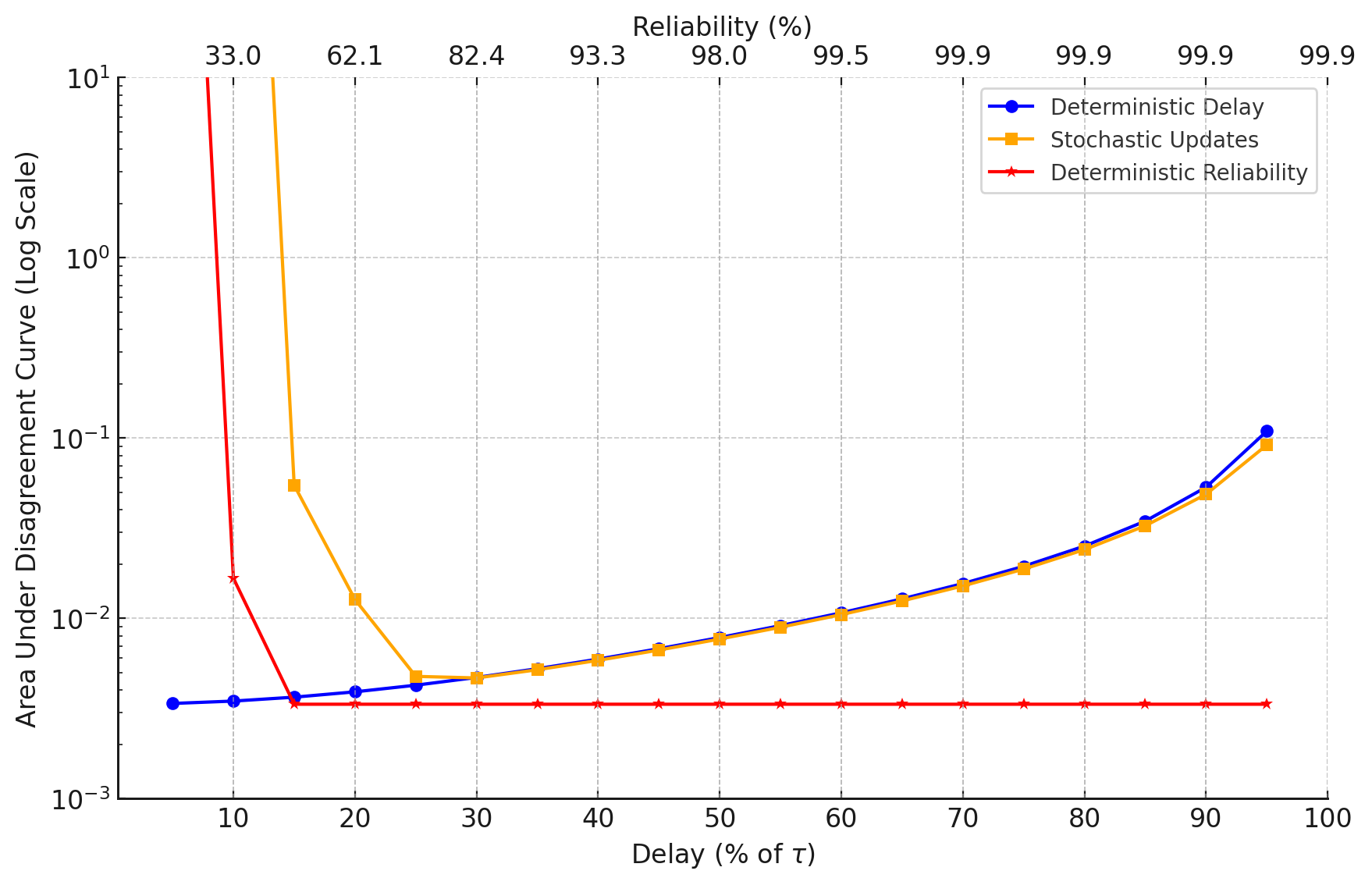}
    \caption{AUC variation vs delay and reliability}
    \label{fig:QoC_tradeoff}
\end{figure}

In Fig. \ref{fig:QoC_tradeoff} we plot how the simulated AUC for the disagreement curves varies as indicated by Eq. \ref{eq:AuC_defn}, plotting $\frac{1}{N} \sum_i J_{\text{norm}_i}^C$, under varying transmission delays and their associated reliability measures. 
Here, we analyze the different curves individually.
The deterministic delay curve shows how the AUC increases as delays rise, assuming no packet drops. As delays grow, the system relies on outdated state information for control inputs, leading to a slower response in adjusting its trajectory. This delay reduces the system’s ability to quickly correct deviations, causing it to take longer to reach consensus.

The reliability curve shows how the AUC changes as system reliability increases, with no delays. Initially, the AUC is high due to packet drops causing outdated information to be used for control commands. As reliability improves, fewer packets are dropped, leading to fresher control inputs, and the AUC gradually converges to its minimum.

The stochastic curve shows the combined effect of both delays and reliability on the AUC. Initially, high AUC values occur due to both packet delays, and drops caused by poor reliability. As delays increase, the curve gradually converges to the deterministic update curve, since higher delays lead to near 100\% reliability, making delays the dominant factor.

The primary takeaway is that it may not always be prudent to minimize the end-to-end delay in the system \cite{latency_min}, but also to understand the reliability-based ramifications that may overall lead to better QoC. Relaxing delays can lead to higher reliability in terms of control packet reception success.

\section{Optimization Framework and Evaluations} \label{results}
We want to now illustrate with an example how the QoC abstraction shown in Section \ref{QOC abstraction} could be utilized. Towards this, we want to maximize the QoC for $N$ collaborating robots, trying to achieve consensus by communicating over a 5G network. In order to realize this, we need to allocate resources to these robots while adhering to resource constraints. 
\subsection{Optimization Problem Formulation}
We use a discretized version of the QoC abstraction, computed from simulations, to formulate an integer linear program (ILP). The ILP is NP-hard and scales exponentially with the number of robots. For a given TDD (Time Division Duplexing) pattern, we optimally allocate resources to robots to maximize their QoC. The parameters of the ILP are as follows:
$ \varphi $ represents the number of TDD slots, \text{N}-number of robots, \(R \)- maximum resources per slot, \(U\) \& \(D\)-the uplink and downlink resources needed per robot, \(\text{G}_l\)-QoC values for delay index \(l\) and finally, \(\Delta_l\)-delay value. UL and DL denote uplink and downlink respectively.
\subsection*{Decision Variables}
\begin{itemize}
    \item \(X_{i,j} \in \mathbb{Z}^{+}\) \quad (Uplink resources allocated to robot \(i\) in slot \(j\)) , 
    \item \(Y_{i,j} \in \mathbb{Z}^{+}\) \quad (Downlink resources allocated to robot \(i\) in slot \(j\)) , 
    \item \(\text{Z}_{i,j}^{\text{UL}} \in \{0, 1\}\) \quad (Binary indicator if robot \(i\) is scheduled in uplink slot \(j\))
    \item \(\text{Z}_{i,j}^{\text{DL}} \in \{0, 1\}\) \quad (Binary indicator if robot \(i\) is scheduled in downlink slot \(j\))
    \item \(\text{S}_i^{\text{First-UL}} \in \mathbb{Z}^{+}\) \quad (First uplink slot for robot \(i\))
    \item \(\text{S}_i^{\text{Last-DL}} \in \mathbb{Z}^{+}\) \quad (Last uplink slot for robot \(i\))
    \item \(\text{S}_i^{\text{First-DL}} \in \mathbb{Z}^{+}\) \quad (First downlink slot for robot \(i\))
    \item \(\text{S}_i^{\text{Last-DL}} \in \mathbb{Z}^{+}\) \quad (Last downlink slot for robot \(i\))
    \item \(\text d_i^{\text{E2E}} \in \mathbb{R}^{+}\) \quad (Time difference for robot \(i\) between last downlink and first uplink overall)
    \item \(\theta_{i,l} \in \{0, 1\}\) \quad (Binary indicator for delay selection by robot \(i\) at index \(l\) with L number of delay indices)
    \item \({Q}_i \in \mathbb{R}^{+}\) \quad (QoC experienced by robot \(i\))
\end{itemize}

Based on these variables, we can define the end-to-end loop delay a robot experiences to be the time difference between the last downlink slot that a robot gets, and the first uplink slot overall that any robot gets, as denoted here:
\begin{equation}
d_i^{\text{E2E}} = S_i^{\text{Last-DL}} - \min\left\{S_{i'}^{\text{First-UL}} \mid i'\in\{1,\cdots,N\} \right\} \forall i 
\end{equation}
We can also now define the allocation delay as:
\begin{equation}
d_i^{\text{alloc}} = S_i^{\text{First-DL}} - \max\left\{S_{i'}^{\text{Last-UL}} \mid i'\in\{1,\cdots,N\} \right\} \forall i 
\end{equation}
These delays have been defined as such after following the information loop as described in Section \ref{system model}.
Now, we proceed to the optimization objective, maximizing $Q_i$ as defined in Section \ref{QoC_model}: 
\begin{equation}
\text{Maximize} \quad \sum_{i=1}^{N} {Q}_i
\end{equation}
    
The constraints used are defined as follows:\\
\begin{equation}
\sum_{i=1, j \in \text{UL}}^{\text{N}} X_{i,j} \leq R \quad \& \quad \sum_{i=1, j \in \text{DL}}^{\text{N}} Y_{i,j} \leq R
\label{resoucre_cons}
\end{equation}

\begin{equation}
\sum_{j \in \text{UL}} X_{i,j} = U \quad \& \quad  \sum_{j \in \text{DL}} Y_{i,j} = D \quad \forall i
\label{UL_DL_cons}
\end{equation}

\begin{equation}
X_{i,j} \leq U \cdot \text{Z}_{i,j}^{\text{UL}} \quad \& \quad Y_{i,j} \leq D \cdot \text{Z}_{i,j}^{\text{DL}} \quad \forall i, j
\label{binary_cons1}
\end{equation}
\begin{equation}
\text{Z}_{i,j}^{\text{UL}}  \leq X_{i,j} \quad \& \quad {Z}_{i,j}^{\text{DL}} \leq Y_{i,j} \quad \forall i, j
\label{binary_cons2}
\end{equation}

\begin{equation}
S_{i}^{\text{First-UL}} \leq (j + 1) + M \cdot (1 - \text{Z}_{i,j}^{\text{UL}}) \quad \forall i, j
\label{first_UL_cons1}
\end{equation}
\begin{equation}
S_{i}^{\text{First-UL}} \geq (j + 1) - M \cdot (1 - \text{Z}_{i,j}^{\text{UL}}) \quad \forall i, j
\label{first_UL_cons2}
\end{equation}

\begin{equation}
\sum_{l=1}^{\text{L}} \theta_{i,l} = 1 \quad \forall i
\label{del_bin_cons}
\end{equation}

\begin{equation}
\begin{aligned}
\text d_i^{\text{E2E}} \leq \Delta_l + ( \varphi - \Delta_l) \cdot (1 - \theta_{i,l}) \quad     \forall i, l
\end{aligned}
\label{e2e_del1}
\end{equation}
\begin{equation}
\begin{aligned}
\text d_i^{\text{E2E}} \geq \Delta_l - \Delta_l \cdot (1 - \theta_{i,l}) \quad \forall i, l
\end{aligned}
\label{e2e_del2}
\end{equation}

\begin{equation}
\sum_{l=1}^{\text{L}} \text{G}_l \cdot \theta_{i,l} = {Q}_i \quad \forall i
\label{QoC_cons}
\end{equation}

Constraint \ref{resoucre_cons} deals with the resource allocation per slot, ensuring the resource constraint for each slot \(j\) is never exceeded.
Constraint \ref{UL_DL_cons} ensures that the total uplink and downlink resource needs for each robot \(i\) are met.
Constraints \ref{binary_cons1} and \ref{binary_cons2} link binary variables to the resource allocation, thereby establishing occurrences of uplink and downlink resources.
Constraints \ref{first_UL_cons1} and \ref{first_UL_cons2} ensure that the first uplink for robot \(i\) occurs in the earliest slot where it is scheduled, with a large constant M. The constraints for the last uplink, first downlink, and last downlink are defined in the same form.    

There are different delay binaries available corresponding to the different delays a robot can experience. Constraint \ref{del_bin_cons} ensures that each robot selects exactly one delay index. This happens with: \(\theta_{i,l}\) only being equal to 1 for just one possible delay index \(l\). Constraints \ref{e2e_del1} and \ref{e2e_del2} help calculate the end-to-end delay a robot experiences. A similar process is followed to compute the allocation delay.
Finally, constraint \ref{QoC_cons} maps the QoC values for each delay index \(l\), given by \(\text{G}_l\). This helps ensure that each robot has only one QoC value, mapped out from the delay it experiences.

Our optimization framework paves the way for capacity optimization supporting multiple objective functions to be understood, such as minimizing the end-to-end delay, maximizing the reliability, etc. 
We can test it for different TDD patterns to understand their corresponding capacity implications.

\subsection{Evaluations} \label{sec:evaluation}
\begin{figure}[t]
    \centering
    \includegraphics[width=0.99\linewidth]{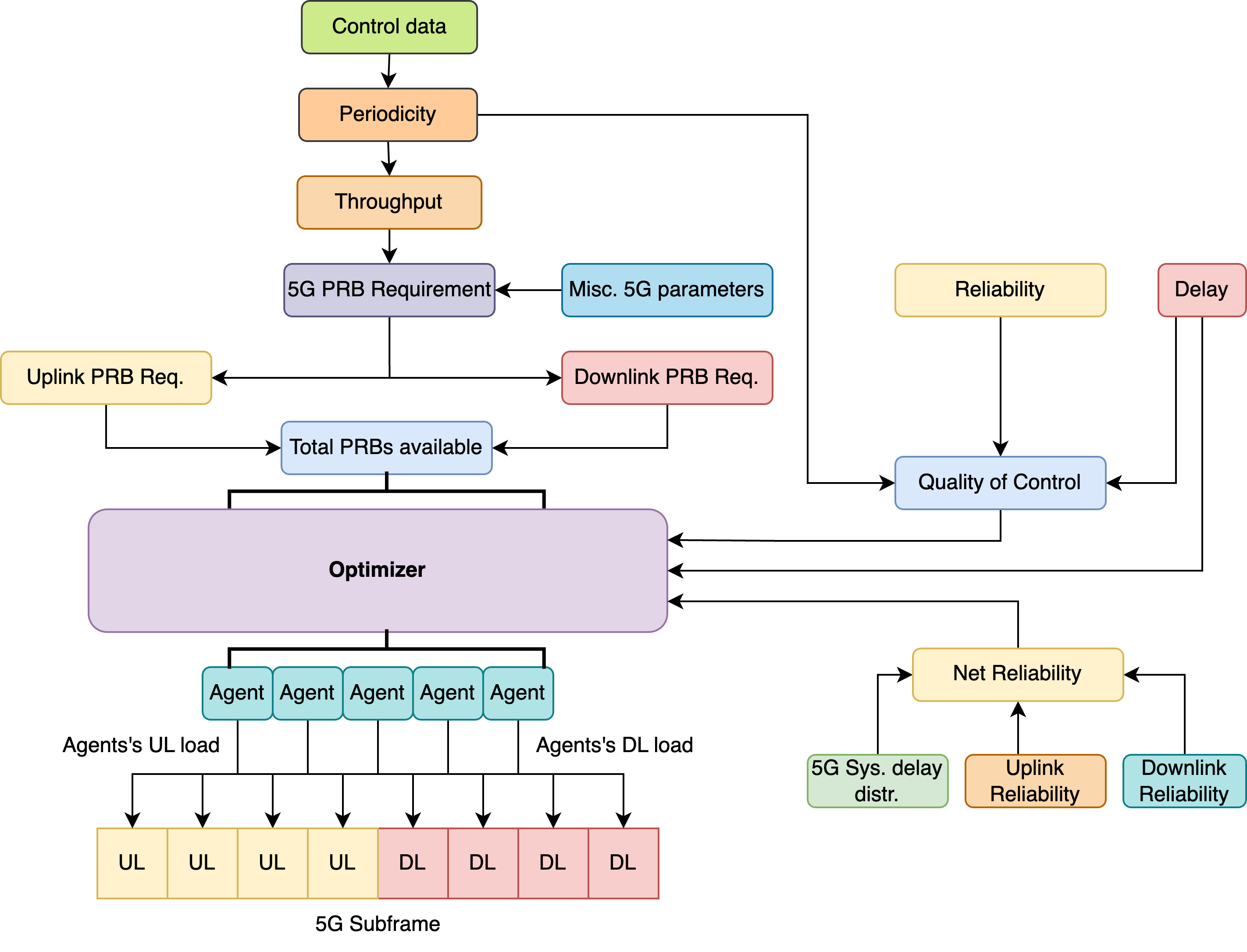}
    \caption{Methodology for our overall optimization framework}
    \label{fig:methodology}
\end{figure}

In this section, we describe how we evaluate our proposed optimization framework, solved using an off-the-shelf solver-Gurobi Optimization.
We feed our optimizer discretized curves (QoC table), derived from curves similar to those shown in Fig. \ref{fig:QoC_tradeoff}, which have been generated with the simulation methodology described in Section \ref{QOC abstraction},  corresponding to all the possible delay values, given a TDD pattern and the net reliability. After obtaining the curves, we proceed towards obtaining the network parameters. As shown in Fig. \ref{fig:full_off}, there is a periodic exchange of information between robots over the 5G network. They can be characterized by UDP-based messages with a packet size of 0.64 kB \cite{s21227463}.
Now, empowered with the transport block to be transmitted, its periodicity, the signal-to-interference-plus-noise ratio (SINR), which is assumed to be constant on average, the corresponding modulation and coding scheme (MCS) index, and finally the PER (obtained with mappings from \cite{Sandra_EESM}), we can calculate the data rate that needs to be delivered. Based on this throughput, as shown in Fig. \ref{fig:methodology}, we can arrive at the uplink and downlink PRB requirements for each robot\cite{etsi_ts_138306_2020}. These robots are then optimally placed by the optimizer in different slots, by looking into their underlying delays, net reliability, and, most importantly, QoC.


The evaluation parameters used, are described in Table \ref{Tab:sim_param}.

\begin{table}[h!]
\begin{center}
\caption{Simulation parameters}
\begin{tabular}{| p{5.5cm} | p{2.5cm} | } \hline
\textbf{Parameters} & \textbf{Chosen values} \\ \hline
Number of aggregated component carriers & $J = 1$ \\ \hline
Scaling factor [3GPP 38.306] & $f = 1$ \\ \hline
Number of MIMO layers & $v = 1$ \\ \hline
Uplink \& Downlink  Overhead & 0.08 \& 0.14\\ \hline
Subframe numerology & $\mu = 1$ \\ \hline
Carrier frequency & FR1 or sub 6 GHz\\ \hline
Maximum number of PRBs available for $\mu = 1$ and BW = 50 MHz & $N_{Max} = 133$ \\ \hline
SINR \& MCS Index & 10dB \& 7 \\ \hline
$PER_{UL}$ \& $PER_{DL}$ & 0.99 \& 0.99 \\ \hline
Number of robots & 80 \\ \hline
Periodicity $\tau$ & 10ms \\ \hline

\end{tabular}
\label{Tab:sim_param}
\end{center}
\end{table}
When the AUC obtained using the curves mentioned above, is greater than 1, potentially signalling instability, we set the QoC for these cases equal to zero. For our evaluations, we compare 4 optimization schemes in total:
\begin{itemize}
    \item Scheme 1 (Proposed), Maximize QoC: The proposed optimization scheme, that takes both the delay and reliability of the communication network into account and maximizes the QoC for the control application. 
    ($\text{Maximize} \quad \sum_{i=1}^{\text{N}} {Q}_i$)
    \item Scheme 2, Minimize delay: This is inspired from \cite{latency_min} and focuses on minimizing the end-to-end delay
    ($\text{Minimize} \quad \sum_{i=1}^{\text{N}} d_i^{\text{E2E}}$)
    \item Scheme 3, Maximize reliability: This scheme maximizes the delay in the system, in turn ensuring the maximum possible reliability for any core and computational delay distribution.
    ($\text{Maximize} \quad \sum_{i=1}^{\text{N}} d_i^{\text{E2E}}$)
    \item Scheme 4 (SOTA), Minimize delay (QoC $>$ 0): This scheme minimizes the maximum delay much like our second evaluation scheme \cite{latency_min}, but only considers those delay values, where the system is stable.
    ($\text{Minimize} \quad \sum_{i=1}^{\text{N}} d_i^{\text{E2E}} \quad \text{subject to} \quad {Q}_i > 0$)
\end{itemize}

\subsubsection{Impact of Delay Distribution}
We evaluate our system using two Gaussian delay distributions, both with a mean of 0.5ms but standard deviations of 0.25ms and 1ms (subsequently referred to as case d1 and d2, respectively), and normalized and clipped to fit the 0-10ms delay range.

\begin{figure}[t]
    \centering
    \includegraphics[width=0.99\linewidth]{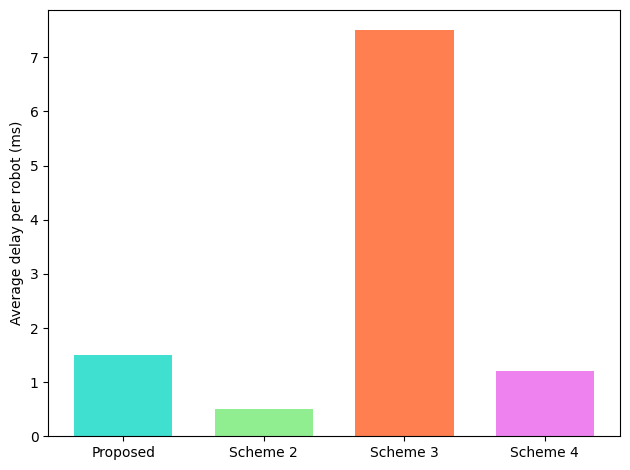}
    \caption{Average delay experienced per robot across different optimization objectives}
    \label{fig:delay_variation}
\end{figure}

\begin{figure}[t]
    \centering
    \includegraphics[width=0.99\linewidth]{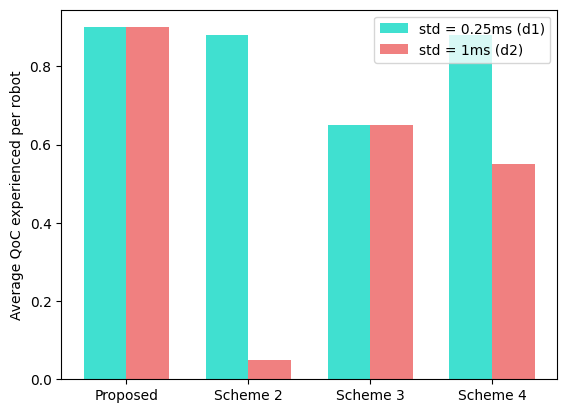}
    \caption{QoC variation for objectives evaluated at different standard deviations}
    \label{fig:QoC_variation}
\end{figure}

Fig. \ref{fig:QoC_variation} shows the average QoC for each robot for case d1 and d2. For d1, all schemes except scheme 3 perform similarly due to high reliability at low delays. Schemes 2 and 4 yield the same results, as no instability arises from the disagreement curves, and schemes 1 and 2 also perform equally well due to high reliability and minimal packet drops. Scheme 3, which maximizes delay, results in a lower QoC due to higher delays.
For d2, lower delay values have low reliability, allowing the proposed optimization scheme (scheme 1) to significantly outperform others by maximizing QoC through balancing delays and reliability. Scheme 2 performs poorly as it selects minimum delays without considering reliability, leading to packet drops. Scheme 4 minimizes delays while factoring in reliability, selecting only values with a positive QoC. Scheme 3, focused on maximizing delay, remains unchanged.
Fig. \ref{fig:delay_variation} shows the average delay experienced by a robot for case d2. Scheme 3 results in the highest delay, while schemes 2 and 4 have the lowest. However, by slightly relaxing the delay, our proposed scheme 1 can achieve a better QoC.

\subsubsection{Impact of TDD Patterns}
\begin{figure}[t]
    \centering
    \includegraphics[width=0.99\linewidth]{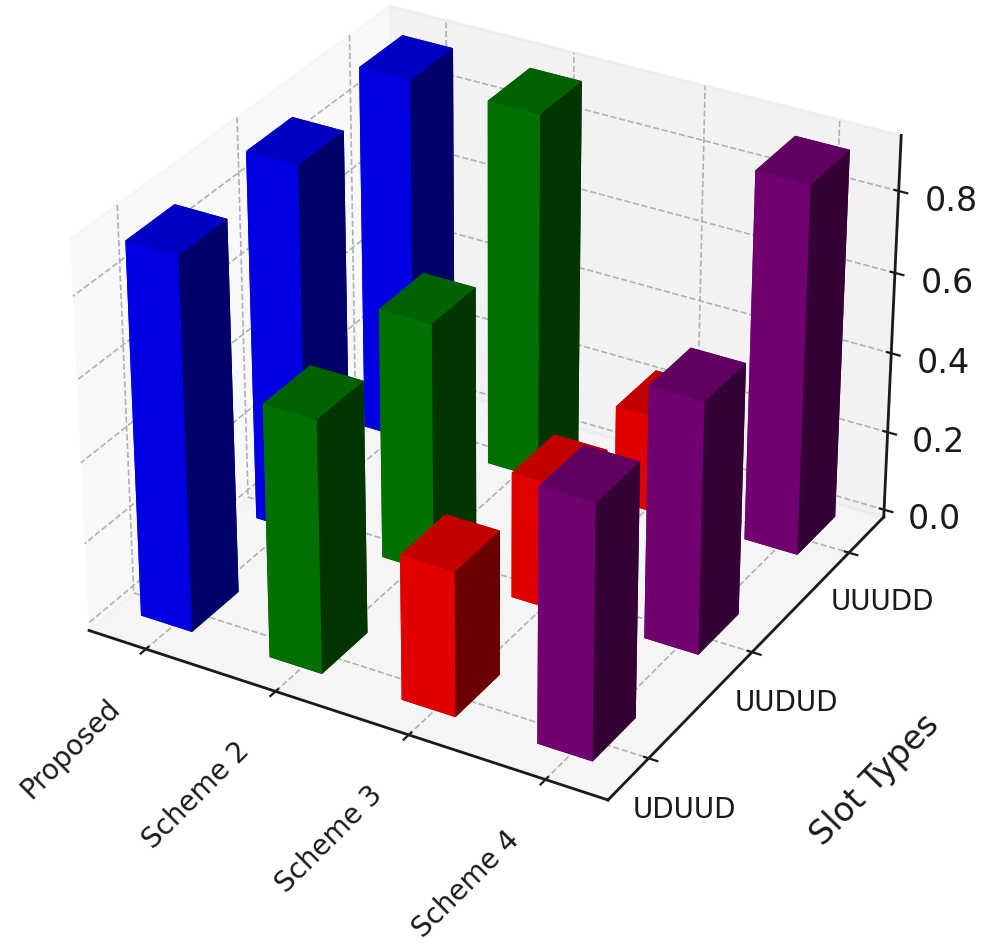}
    \caption{QoC variation for different objectives vs TDD patterns}
    \label{fig:3dbar_TDD}
\end{figure}

In Fig. \ref{fig:3dbar_TDD} shows how QoC per robot varies with different TDD patterns, given the same distribution and number of uplink and downlink slots. We evaluate three patterns: UDUUD, UUDUD, and UUUDD, where U and D represent uplink and downlink slots. These patterns are repeated over the 5G NR frame. As seen in Fig. \ref{fig:3dbar_TDD}, we can observe that our proposed framework is more advantageous for UUDUD and UDUUD. This happens since the closer we allocate the uplink and downlink slots in a TDD pattern, the more likely it is for the system to drop packets at lower delays and reliability (in case of Scheme 2 and 4), if it is agnostic of packet reliability. Scheme 1 relaxes the delays again, allowing more control packets to be delivered reliably, leading to better average QoC per robot. 
When the uplink and downlink slots are placed further apart as in the case with UUUDD, scheme 1 performs similarly to schemes 2 and 4. Scheme 1, which relaxes delays for better reliability, consistently achieves the highest QoC across patterns. 
\section{Conclusion} \label{conclusion}
In this work, we provided a Quality of Control (QoC) based abstraction of robotic workloads, parameterized by network delays and reliability in a 5G network. We exemplified our approach using consensus in collaborative robotics and solved an ILP to optimally allocate network resources to different robots, while maximizing their QoC, given a TDD pattern. We show that it may not always be judicious to minimize the end-to-end delay in the system purely, but to sometimes relax that delay allowing for control packets to be delivered more reliably. Our results indicate substantial gains in terms of maximizing the QoC for robots, which translates to almost 32\% less energy being spent for some applications compared to SOTA, towards convergence to the consensus objective. Our methodology extends to different objective functions and parameters such as number of robots, periodicity, etc. as well.

\bibliographystyle{ieeetr}
\bibliography{references}

\begin{thebibliography}{10}

\bibitem{5gsmart2020}
L.~G. et~al., ``5{G}-smart deliverable d1.1 - smart manufacturing in the context of 5{G},'' tech. rep., 2020.

\bibitem{7883994}
M.~Wollschlaeger, T.~Sauter, and J.~Jasperneite, ``The future of industrial communication: Automation networks in the era of the internet of things and industry 4.0,'' {\em IEEE Industrial Electronics Magazine}, vol.~11, no.~1, pp.~17--27, 2017.

\bibitem{edge_robotics}
A.~Baxi, M.~Eisen, S.~Sudhakaran, F.~Oboril, G.~S. Murthy, V.~S. Mageshkumar, M.~Paulitsch, and M.~Huang, ``Towards factory-scale edge robotic systems: Challenges and research directions,'' {\em IEEE Internet of Things Magazine}, vol.~5, no.~3, pp.~26--31, 2022.

\bibitem{mesbahi2010graph}
M.~Mesbahi and M.~Egerstedt, {\em Graph Theoretic Methods in Multiagent Networks}, vol.~33.
\newblock Princeton University Press, 2010.

\bibitem{QoC_Assessment}
K.~Polachan, J.~Pa, C.~Singh, and T.~Prabhakar, ``Assessing quality of control in tactile cyber--physical systems,'' {\em IEEE Transactions on Network and Service Management}, vol.~19, no.~4, pp.~5348--5365, 2022.

\bibitem{latency_min}
A.~A. Sardar, A.~S. Rao, T.~Alpcan, G.~Das, and M.~Palaniswami, ``Network resource allocation for industry 4.0 with delay and safety constraints,'' {\em IEEE Transactions on Cognitive Communications and Networking}, vol.~10, no.~1, pp.~223--237, 2024.

\bibitem{collab4g}
A.~Romero, C.~Delgado, L.~Zanzi, R.~Suárez, and X.~Costa-Pérez, ``Cellular-enabled collaborative robots planning and operations for search-and-rescue scenarios,'' in {\em 2024 ICRA}, 2024.

\bibitem{WifiEdge}
S.~Mohanti, D.~Roy, M.~Eisen, D.~Cavalcanti, and K.~Chowdhury, ``{L-NORM: Learning and Network Orchestration at the Edge for Robot Connectivity and Mobility in Factory Floor Environments},'' {\em IEEE Transactions on Mobile Computing}, vol.~23, no.~4, pp.~2898--2914, 2024.

\bibitem{lat-rel}
K.~Gatsis, H.~Hassani, and G.~J. Pappas, ``Latency-reliability tradeoffs for state estimation,'' {\em IEEE Transactions on Automatic Control}, vol.~66, no.~3, pp.~1009--1023, 2021.

\bibitem{cao2012overview}
Y.~Cao, W.~Yu, W.~Ren, and G.~Chen, ``An overview of recent progress in the study of distributed multi-agent coordination,'' {\em IEEE Transactions on Industrial informatics}, vol.~9, no.~1, pp.~427--438, 2012.

\bibitem{s21227463}
P.~Obal and P.~Gierlak, ``{EGM} toolbox—interface for controlling {ABB} robots in simulink,'' {\em Sensors}, vol.~21, no.~22, 2021.

\bibitem{10558844}
H.~Lyu, J.~Yan, J.~Zhang, Z.~Pang, G.~Yang, and A.~J. Isaksson, ``Cloud–fog automation,'' {\em IEEE Industrial Electronics Magazine}, 2024.

\bibitem{QoC_energy_2}
A.~Aminifar, P.~Eles, Z.~Peng, A.~Cervin, and K.-E. Årzén, ``Control-quality-driven design of embedded control systems with stability guarantees,'' {\em IEEE Design \& Test}, vol.~35, no.~4, pp.~38--46, 2018.

\bibitem{Sandra_EESM}
S.~Lagen, K.~Wanuga, H.~Elkotby, S.~Goyal, N.~Patriciello, and L.~Giupponi, ``New radio physical layer abstraction for system-level simulations of 5{G} networks,'' in {\em ICC 2020}, pp.~1--7, 2020.

\bibitem{etsi_ts_138306_2020}
ETSI, ``5{G} {NR} user equipment ({UE}) radio access capabilities {3GPP} rel. 17,'' tech. rep., 2022.

\end{thebibliography}
\vspace{12pt}

\end{document}